\documentclass{article}
\usepackage{ijcai26}

\usepackage{times}
\usepackage{soul}
\usepackage{url}
\usepackage[hidelinks]{hyperref}
\usepackage[utf8]{inputenc}
\usepackage[small]{caption}
\usepackage{graphicx}
\usepackage{amsmath}
\usepackage{amsthm}
\usepackage{booktabs}
\usepackage{algorithm}
\usepackage{algorithmic}
\usepackage{multirow}
\usepackage[switch]{lineno}
\usepackage{amssymb}
\usepackage{tabularx}

\usepackage[most]{tcolorbox}
\usepackage{xcolor}

\tcbset{
  promptouter/.style={
    boxrule=0.8pt,
    colframe=black,
    colback=white,
    arc=2mm,
    left=6pt,
    right=6pt,
    top=6pt,
    bottom=6pt,
    enhanced,
    breakable
  },
  promptinner/.style={
    boxrule=0pt,
    arc=1.5mm,
    left=6pt,
    right=6pt,
    top=4pt,
    bottom=4pt,
    enhanced,
    breakable
  }
}

\title{Probing Prompt Design for Socially Compliant Robot Navigation with Vision Language Models}

\author{
Ling Xiao$^1$
\and
Toshihiko Yamasaki$^2$
\affiliations
$^1$Hokkaido University, 9-9 Kita-14-jo Nishi, Kita-ku, Sapporo, Hokkaido 060-0814, Japan\\
$^2$The University of Tokyo, 7-3-1 Hongo, Bunkyo-ku, Tokyo 113-8656, Japan\\
\emails
ling@ist.hokudai.ac.jp,
yamasaki@cvm.t.u-tokyo.ac.jp
}

\begin{document}

\maketitle

\begin{abstract}
Language models are increasingly used for social robot navigation, yet existing benchmarks largely overlook principled prompt design for socially compliant behavior. This limitation is particularly relevant in practice, as many systems rely on small vision language models (VLMs) for efficiency. Compared to large language models, small VLMs exhibit weaker decision-making capabilities, making effective prompt design critical for accurate navigation.
Inspired by cognitive theories of human learning and motivation, we study prompt design along two dimensions: system guidance (action-focused, reasoning-oriented, and perception–reasoning prompts) and motivational framing, where models compete against humans, other AI systems, or their past selves.
Experiments on two socially compliant navigation datasets reveal three key findings. First, for non-finetuned GPT-4o, competition against humans achieves the best performance, while competition against other AI systems performs worst. For finetuned models, competition against the model’s past self yields the strongest results, followed by competition against humans, with performance further influenced by coupling effects among prompt design, model choice, and dataset characteristics. Second, inappropriate system prompt design can significantly degrade performance, even compared to direct finetuning. Third, while direct finetuning substantially improves semantic-level metrics such as perception, prediction, and reasoning, it yields limited gains in action accuracy. In contrast, our system prompts produce a disproportionately larger improvement in action accuracy, indicating that the proposed prompt design primarily acts as a decision-level constraint rather than a representational enhancement.
\end{abstract}

\begin{figure}[t]
    \centering
    \includegraphics[width=\linewidth]{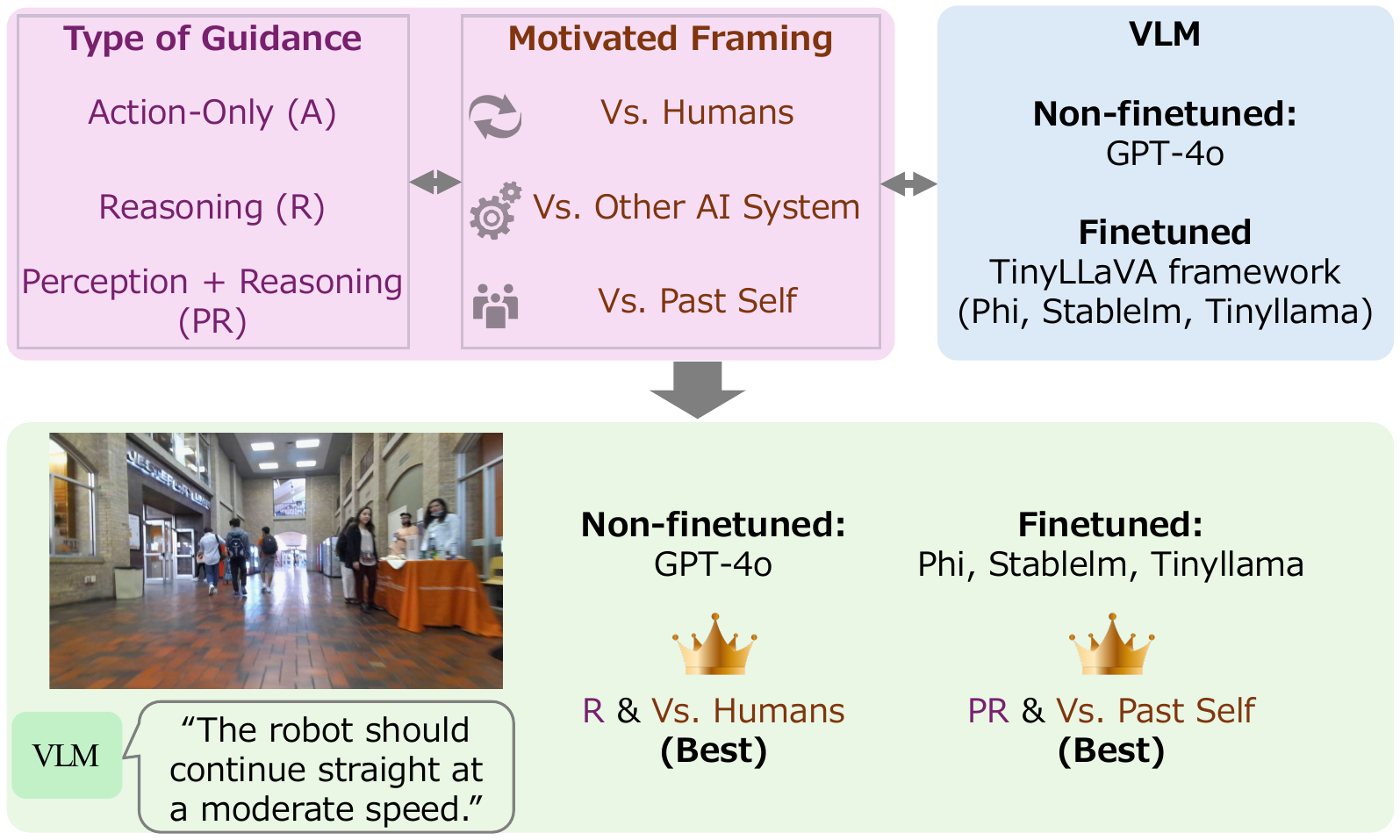}
    \caption{Non-finetuned GPT-4o performs best with reasoning-focused prompts and human-competition framing, while finetuned VLMs benefit most from perception–reasoning prompts combined with competition against the model’s past behavior.}
    \label{fig:teaser}
\end{figure}

\section{Introduction}
\label{sec:intro}

Mobile robots are increasingly deployed in a wide range of real-world applications, including healthcare and eldercare assistance, delivery and logistics, as well as security and surveillance. Many of these applications require robots to operate in public spaces shared with pedestrians, where they must interact naturally with humans and navigate safely in dynamic, socially complex environments. In such settings, it is essential for robots to exhibit socially compliant behaviors in both interaction and navigation, ensuring not only physical safety but also efficiency, comfort, and user acceptance~\cite{francis2025principles,payandeh2024social}. First, robots must accurately perceive their surroundings, understand and anticipate human intentions, and reason under uncertainty in dynamic and cluttered environments. Moreover, they must balance navigation efficiency with social norms, safety constraints, and human comfort. Addressing these challenges requires tightly integrated perception, prediction, and planning capabilities that can adapt robot behavior to diverse and evolving interaction scenarios.

Existing approaches to socially compliant navigation can be broadly categorized into imitation learning (IL)-based methods~\cite{cuan2024gesture2path}, reinforcement learning (RL)-based methods~\cite{kathuria2025learning}, and vision language model (VLM)-based approaches~\cite{song2024vlm}. Among these approaches, VLM-based methods have recently attracted increasing attention due to the strong contextual understanding and commonsense reasoning capabilities of modern language models, as well as their inherent explainability. For example, VLM-Social-Nav~\cite{song2024vlm} leverages GPT-4V to generate navigation instructions conditioned on visual observations. Kawabata et al.~\cite{kawabata2025socialnav} propose SocialNav-MoE, an efficient mixture-of-experts VLM for socially compliant navigation trained with reinforcement fine-tuning. Wang et al.~\cite{wang2025maction} introduce MAction-SocialNav, which explicitly addresses action ambiguity by enabling the generation of multiple plausible actions within a single scenario. Kong et al.~\cite{kong2025autospatial} present AutoSpatial, a model designed for efficient spatial reasoning in social robot navigation. Narasimhan et al.~\cite{narasimhan2025olivia} propose OLiVia-Nav, which incorporates both social and environmental context during robot trajectory planning and adapts to previously unseen social scenarios. Liu et al.~\cite{liu2025muson} propose the MUSON dataset for this task and evaluate the performance of several small VLMs on it.

Despite these advances, existing studies primarily focus on model architectures and training strategies, while prompt design remains largely underexplored. This limitation becomes particularly critical in efficiency-oriented deployments that rely on small VLMs~\cite{wang2025maction}. In such settings, well-designed prompts play a crucial role in stimulating the decision-making capability, thereby compensating for the limited capacity of smaller models.

In this paper, we address this gap by systematically investigating prompt design for socially compliant robot navigation (as shown in Fig.~\ref{fig:teaser}). Inspired by theories of human cognitive learning, we study how different types of guidance, including action focused prompts, reasoning oriented prompts, and prompts that integrate perception and reasoning, as well as different motivational framing strategies (competing against humans, other AI models, or the model’s past self), influence navigation performance.
We conduct experiments on two benchmark datasets and show that prompt design plays a critical role in shaping model behavior. Our results demonstrate that well designed prompts can substantially improve decision quality and social compliance, particularly in resource constrained settings based on small VLMs. Overall, the main contributions of this work are summarized as follows:
\begin{itemize}
    \item We conduct a systematic study of prompt design for socially compliant navigation, focusing on system guidance and motivational framing.
    
    \item We show that optimal prompt configurations differ across models: reasoning-oriented prompts with competition against humans perform best for non-finetuned GPT-4o, while perception–reasoning prompts combined with competition against the model’s past behavior are most effective for finetuned small VLMs, with performance influenced by coupling effects among prompts, models, and datasets.
    
    \item We identify a semantic–decision mismatch: finetuning mainly improves semantic-level metrics, whereas our proposed system prompts yield larger gains in action accuracy, indicating that prompt design primarily functions as a decision-level constraint.
\end{itemize}

\section{Related Work}
\subsection{Social Robot Navigation}
For social robot navigation, safety is paramount~\cite{liang2021crowd}. Classical methods enforce collision constraints or fuse multi-sensor data (2-D LiDAR, depth cameras) for smooth avoidance~\cite{liang2021crowd}.
Safety alone, however, is insufficient in human-populated spaces. Robots must also respect social norms (such as personal space, group dynamics, cultural conventions) to be perceived as acceptable and trustworthy. Traditional methods often ignore such social norms, treating pedestrians merely as dynamic obstacles.

Learning-based approaches seek to encode both safety and social awareness. Demonstration-driven motion learning~\cite{sun2021motion} and Reinforcement Learning (RL) in simulators~\cite{liang2021crowd} show promise but depend on large datasets or highly realistic human simulations, which rarely capture nuanced interactions, yielding policies with poor generalization.

Recently, VLMs have opened new directions. Several VLMs are developed to generate high-level actions~\cite{payandeh2024social,liu2025muson,wang2025maction,kawabata2025socialnav}, evaluate trajectories~\cite{narasimhan2024olivia}, and predict directions~\cite{song2024vlm}. Datasets such as SCAND~\cite{karnan2022socially}, MuSoHu~\cite{nguyen2023toward}, SNEI~\cite{payandeh2024social}, and MUSON~\cite{liu2025muson} have further advanced research in this domain.

\subsection{Language Prompt Tuning}
Language prompt tuning aims to improve the practical applicability of large-scale pre-trained language models by reformulating downstream tasks into natural language instructions that align with the models’ pretraining objectives~\cite{brown2020language,zhou2022learning}. In this paradigm, task performance critically depends on how the task is expressed through a prompt template and a verbalizer, which together define the interface between the model and the target task.

A variety of prompt-based learning strategies have been explored, including handcrafted prompts~\cite{xiao2025llm,schick2021s}, prompt mining and paraphrasing~\cite{jiang2020can}, gradient-based prompt search~\cite{shin2020autoprompt}, and automatic prompt generation~\cite{gao2021making}. While these approaches demonstrate the effectiveness of prompt-based learning, prior studies have shown that discrete prompts are often highly sensitive to wording and formatting, which can lead to unstable performance across different tasks and models~\cite{zhao2021calibrate,liu2024gpt}.

This sensitivity is particularly problematic in decision-critical and embodied scenarios such as socially compliant navigation, where many existing methods rely on small VLMs to achieve higher efficiency. However, small VLMs typically exhibit weaker decision-making capabilities than large models, making their behavior more vulnerable to suboptimal prompt formulations. As a result, how to retain computational efficiency while improving decision accuracy remains an open challenge. In this paper, we address this challenge through principled system prompt design.

\section{Proposed Methods}
\subsection{Proposed Prompts}
Inspired by theories of human cognitive learning and behavioral motivation~\cite{dimenichi2015power}, which suggest that human decision making is strongly influenced by both instructional structure and competitive incentives, we investigate prompt design for socially compliant robot navigation along two coupled dimensions: system guidance and motivational framing. Our goal is to understand how different combinations of these two dimensions affect the decision-making behavior of small VLMs, particularly in efficiency-oriented navigation settings.

Specifically, we design a set of textual system prompts that vary in the type of guidance provided to the model. These include action-focused prompts that directly constrain the model’s output to navigation actions, reasoning-oriented prompts that encourage explicit deliberation before action selection, and prompts that integrate perception and reasoning to jointly model environmental understanding and decision making. In parallel, we introduce different motivational framing strategies, where the model is instructed to perform competitively against humans, other AI systems, or its own past behavior. By systematically combining different guidance types with distinct motivational signals, we construct a prompt catalog that enables controlled analysis of their individual and joint effects on navigation performance. This design allows us to examine not only which prompt configurations yield higher action accuracy, but also how prompt structure influences the alignment between perception, reasoning, and final action decisions. Detailed prompt formulations used in our experiments are summarized below.

\begin{tcolorbox}[
  promptouter,
  title={Textual System Prompt Catalog},
  width=\columnwidth,
  boxsep=1mm,
  left=1mm,
  right=1mm,
]

\small
\textbf{Shared Base Prompt.} You are an intelligent assistant specializing in socially compliant robot navigation. You must understand human behaviors, infer intentions, and plan safe, smooth, and socially appropriate paths.

\medskip
\textbf{Prompt Variants.}

\setlength{\tabcolsep}{4pt}
\renewcommand{\arraystretch}{1.05}
\begin{tabularx}{\linewidth}{@{}lX@{}}
\toprule
\textbf{ID} & \textbf{Additional Instruction} \\
\midrule
A1  & Perform competitively against humans. \\
A2  & Perform competitively against other AI systems. \\
A3  & Perform competitively against your past self. \\
R1  & Explain your reasoning clearly and perform competitively against humans. \\
R2  & Explain your reasoning clearly and perform competitively against other AI systems. \\
R3  & Explain your reasoning clearly and perform competitively against your past self. \\
PR1 & Explain your perception and reasoning clearly while performing competitively against humans. \\
PR2 & Explain your perception and reasoning clearly while performing competitively against other AI systems. \\
PR3 & Explain your perception and reasoning clearly while performing competitively against your past self. \\
\bottomrule
\end{tabularx}

\end{tcolorbox}

\subsection{Model Structure}

\begin{figure}[t]
    \centering
    \includegraphics[width=0.8\linewidth]{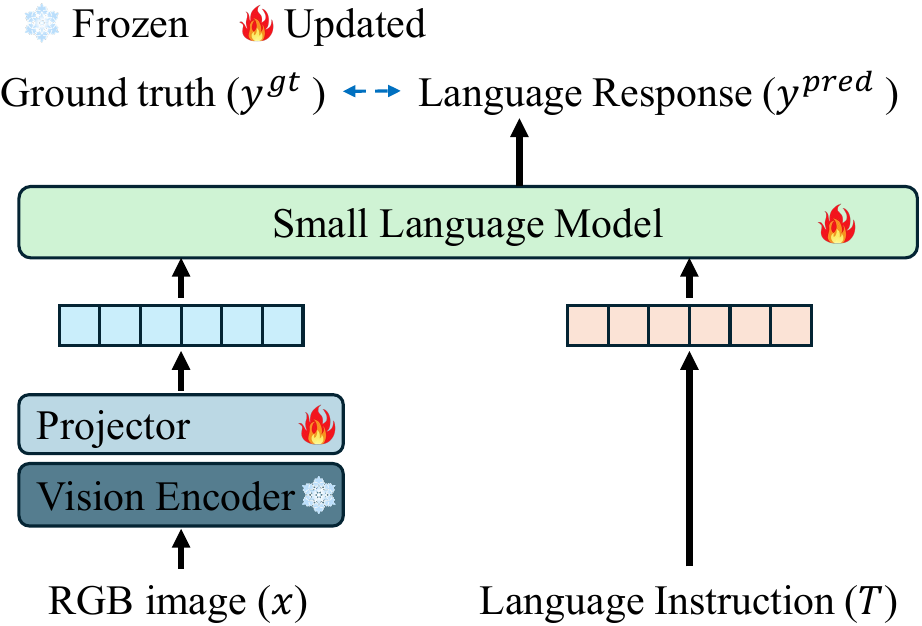}
    \caption{Overview of the TinyLLaVA framework used in our experiments.}
    \label{fig:tinyllava}
\end{figure}

We adopt the TinyLLaVA framework as the VLM architecture in our study.
As illustrated in Fig.~\ref{fig:tinyllava}, the model consists of a frozen vision encoder, a trainable multimodal projector, and a trainable small language model.
The vision encoder extracts visual features from the input RGB image and remains frozen during training.
These features are then mapped into the language embedding space through the projector, which is updated to enable effective cross modal alignment.
The projected visual tokens, together with the language instruction tokens, are fed into the small language model to generate task specific language responses.

\section{Experimental Setup}
\subsection{Implementation Details}

All experiments are conducted using the TinyLLaVA framework with distributed training enabled by DeepSpeed.
We fine-tune all VLMs on four NVIDIA GPUs using DeepSpeed ZeRO Stage-3 for memory-efficient optimization.
The language model is fully fine-tuned, while the vision tower remains frozen throughout training.
The multimodal connector is trained end-to-end to enable effective cross-modal alignment.
SigLIP~\cite{zhai2023sigmoid} is adopted as the vision encoder.
FlashAttention~2 is used to accelerate attention computation, and all models are trained in FP16 precision.
Models are trained for 5 epochs with a per-device batch size of 2 and gradient accumulation of 4 steps,
resulting in an effective batch size of 32.
We use the AdamW optimizer with an initial learning rate of $2\times10^{-5}$,
a cosine learning rate schedule with a warmup ratio of 3\%, and no weight decay.
The number of data loader workers is set to 8, without modality-length-based sample grouping.
\begin{figure*}[t]
    \centering
    \includegraphics[width=\linewidth]{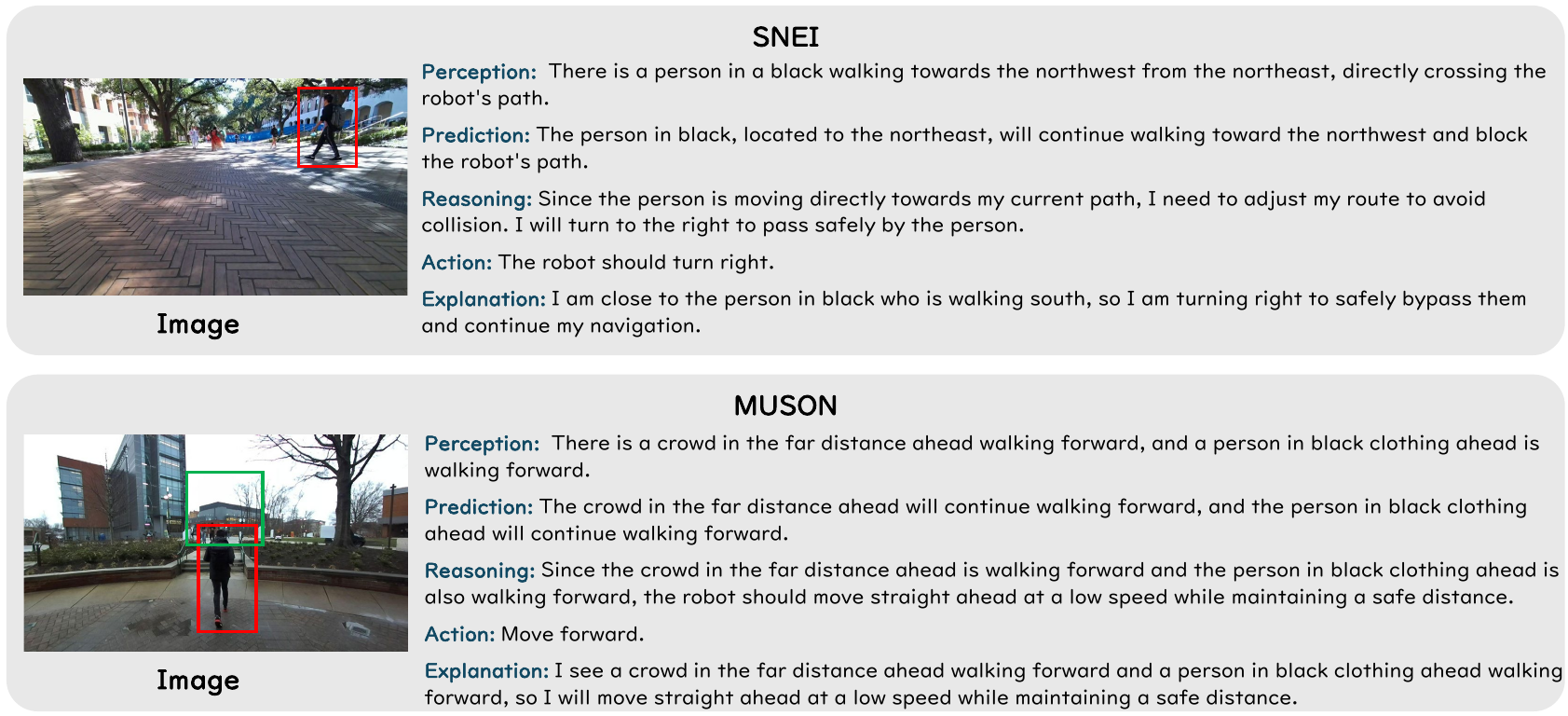}
    \caption{Visualizations of the samples from SNEI and MUSON datasets.}
    \label{fig:dataset}
\end{figure*}
\subsection{Datasets.}
We evaluate our method on SNEI~\cite{payandeh2024social} and MUSON~\cite{liu2025muson} datasets. Fig.~\ref{fig:dataset} presents representative samples from the datasets. Specifically, SNEI contains 325 egocentric images with five-turn conversational annotations provided by four independent annotators. In our experiments, 265 samples are used for training and 60 samples are reserved for testing. MUSON consists of 800 curated egocentric samples that explicitly model both dynamic and static constraints. The dataset is annotated using a five-step chain of thought pipeline with a human and VLM and human double-checking procedure. We use 640 samples for training and 160 samples for testing.
Both datasets cover diverse indoor and outdoor environments with varying crowd densities and two different robot platforms, which helps mitigate the risk of overfitting to narrow annotator bias.

\begin{table}[t]
\centering
\caption{Performance in terms of Action Accuracy (AA). Higher values indicate better action-level decision accuracy. The best-performing result is highlighted in bold. AA$_{\text{SNEI}}$ denotes AA on the SNEI dataset, and AA$_{\text{MUSON}}$ denotes AA on the MUSON dataset.}
\label{tab:gpt-4o}
\resizebox{0.5\textwidth}{!}{
\begin{tabular}{l l l l c c}
\toprule
LLM & System Prompt & Prompt Type & Variant & AA$_{\text{SNEI}}$ $\uparrow$ &
AA$_{\text{MUSON}}$ $\uparrow$ \\
\midrule
\multirow{11}{*}{GPT-4o}
& No System Prompt & -- & --  & 0.550 & 0.531\\
\cline{2-6}
& \multirow{10}{*}{Textual System Prompts}
& -- & Raw & 0.500 & 0.500\\
\cline{3-6}
&  & \multirow{3}{*}{A-Series} & A1 & 0.600 & 0.531\\
&  &  & A2 & 0.500 & 0.531 \\
&  &  & A3 & 0.517  & 0.531\\
\cline{3-6}
&  & \multirow{3}{*}{R-Series} & R1 & \textbf{0.633}  &\textbf{0.563} \\
&  &  & R2 & 0.550 & 0.538 \\
&  &  & R3 & 0.550  & 0.544 \\
\cline{3-6}
&  & \multirow{3}{*}{PR-Series} & PR1 & 0.550 & 0.550\\
&  &  & PR2 & 0.550 &0.544 \\
&  &  & PR3 & 0.583 &0.550 \\
\bottomrule
\end{tabular}
}
\end{table}

\begin{table*}[htbp]
\centering
\caption{Performance on the SNEI dataset. We evaluate action accuracy and semantic similarity for perception, prediction, and reasoning.}
\label{tab:SNEI_AA_Semantic}
\resizebox{0.98\textwidth}{!}{
\begin{tabular}{l l l l c cc cc cc}
\toprule
\multirow{3}{*}{LLM} 
& \multicolumn{3}{c}{Prompt} 
& \multirow{3}{*}{AA ($\uparrow$)} 
& \multicolumn{6}{c}{Semantic Similarity ($\uparrow$)} \\
\cmidrule(lr){2-4} \cmidrule(lr){6-11}
& \multirow{2}{*}{Prompt Type} & \multirow{2}{*}{Prompt Family} & \multirow{2}{*}{Variant} 
&  
& \multicolumn{2}{c}{Perception (Q1)}
& \multicolumn{2}{c}{Prediction (Q2)}
& \multicolumn{2}{c}{Reasoning (Q3)} \\
\cmidrule(lr){6-7} \cmidrule(lr){8-9} \cmidrule(lr){10-11}
&  &  & 
&  
& BERT-F1 & SBERT
& BERT-F1 & SBERT
& BERT-F1 & SBERT \\
\midrule

\multirow{11}{*}{Phi-2-2.7B~\cite{javaheripi2023phi}}
& No System Prompt & -- & -- & 0.450 & 0.349 &  0.504 & 0.386 & 0.590 &0.324  & 0.707 \\
\cline{2-11}
& \multirow{10}{*}{Textual System Prompts} & -- & Raw & 0.433 & 0.354 & 0.593 & 0.346 &0.419  & 0.298 & 0.623 \\
\cline{3-11}
&  & \multirow{3}{*}{A-Series} & A1 & 0.267 & 0.358 & 0.615 &0.295  & 0.335 &  0.313 &0.652  \\
&  &  & A2 & 0.267 & 0.349 & 0.610 &0.289  & 0.341 &  0.315 &0.648  \\
&  &  & A3 & \textbf{0.317} & 0.387 & 0.654 &0.332  & 0.362 &  0.341 &0.679  \\
\cline{3-11}
&  & \multirow{3}{*}{R-Series} & R1& 0.300 & 0.364 & 0.632 &0.309  & 0.341 &  0.339 &0.664  \\
&  &  & R2 & 0.267 & 0.344 & 0.611 &0.275  & 0.352 &  0.317 &0.651  \\
&  &  & R3 & \textbf{0.350} & 0.393 & 0.664 &0.341  & 0.373 &  0.355 &0.683 \\
\cline{3-11}
&  & \multirow{3}{*}{PR-Series} & PR1 &  0.383 &  0.357 & 0.620 &  0.317 & 0.374  & 0.326  &0.620   \\
&  &  & PR2 & 0.317 &  0.352 &0.628  & 0.412 & 0.471 & 0.373 & 0.669 \\
&  &  & PR3 & \textbf{0.500} &  0.376 &0.668  & 0.500  & 0.630 & 0.402 & 0.734 \\
\midrule

\multirow{11}{*}{Stablelm-2-zephyr-1\_6b~\cite{bellagente2024stable}}
& No System Prompt & -- & -- &0.483 & 0.345 & 0.633 &  0.372 &  0.522& 0.354 & 0.660\\
\cline{2-11}
& \multirow{10}{*}{Textual System Prompts} & -- & Raw &0.467 & 0.327 & 0.632 &  0.359 &  0.510& 0.337 & 0.658\\
\cline{3-11}
&  & \multirow{3}{*}{A-Series} & A1 & 0.450 &0.375  &0.612  &  0.419 &  0.493 &0.297  &0.626 \\
&  &  & A2 &0.367  & 0.373 &0.644 &0.463 &  0.558 & 0.381 & 0.680 \\
&  &  & A3 & \textbf{0.467} & 0.380 &  0.671& 0.453 &  0.527 &0.357  & 0.698 \\
\cline{3-11}
&  & \multirow{3}{*}{R-Series} & R1 & 0.333  & 0.341 & 0.559 & 0.407 & 0.392 & 0.307 & 0.583 \\
&  &  & R2 & 0.317 &  0.380 & 0.626  &0.497  & 0.538 & 0.369 &0.667   \\
&  &  & R3 & \textbf{0.417} & 0.364 & 0.609 & 0.415 & 0.532 & 0.320 &0.628  \\
\cline{3-11}
&  & \multirow{3}{*}{PR-Series} & PR1 & 0.317  & 0.317 & 0.601 & 0.311 & 0.427 & 0.291 & 0.579 \\
&  &  & PR2& 0.333 &  0.338 & 0.632  &0.475  & 0.521 & 0.340 &0.661   \\
&  &  & PR3 & \textbf{0.517} & 0.423 & 0.632 & 0.443 & 0.573 & 0.374 &0.675  \\
\midrule

\multirow{11}{*}{TinyLlama-1.1B-Chat-v1.0~\cite{zhang2024tinyllama}}
& No System Prompt & -- & -- & 0.417 & 0.360 & 0.660 &0.429  &  0.555 & 0.398 & 0.717 \\
\cline{2-11}
& \multirow{10}{*}{Textual System Prompts} & -- & Raw & 0.433 & 0.304 & 0.611 & 0.408 &  0.456 & 0.385 & 0.692\\
\cline{3-11}
&  & \multirow{3}{*}{A-Series} & A1 & 0.383 &0.338 & 0.522 &0.466 &0.506 & 0.382 & 0.662 \\
&  &  & A2 & 0.383 &0.333  &0.583 & 0.376 & 0.465 & 0.380 & 0.666 \\
&  &  & A3 & \textbf{0.450} & 0.336 & 0.596 & 0.400& 0.543& 0.355 & 0.648 \\
\cline{3-11}
&  & \multirow{3}{*}{R-Series} & R1 & 0.467 &0.283 &0.625  &0.375 &0.462 &0.352  & 0.673 \\
&  &  & R2 & 0.333 &  0.367 &  0.611 & 0.404&0.476 &0.381 & 0.688 \\
&  &  & R3 & \textbf{0.483} &0.322  &  0.656 & 0.378 & 0.495 & 0.372 & 0.691\\
\cline{3-11}
&  & \multirow{3}{*}{PR-Series} & PR1 &0.433 & 0.389 & 0.667 & 0.447 &0.496 & 0.395 &0.717  \\
&  &  & PR2 & 0.317 &  0.298&0.430  &0.361  &0.418  & 0.265 &  0.495\\
&  &  & PR3  & \textbf{0.500}  & 0.386 & 0.689 &0.474  & 0.547 & 0.402 & 0.732\\
\bottomrule
\end{tabular}
}
\vspace{0.2cm}
\centering
\caption{Performance on the MUSON dataset. We evaluate action accuracy and semantic similarity for perception, prediction, and reasoning.}
\label{tab:MUSON_AA_Semantic}
\resizebox{0.98\textwidth}{!}{
\begin{tabular}{l l l l c cc cc cc}
\toprule
\multirow{3}{*}{LLM} 
& \multicolumn{3}{c}{Prompt} 
& \multirow{3}{*}{AA ($\uparrow$)} 
& \multicolumn{6}{c}{Semantic Similarity ($\uparrow$)} \\
\cmidrule(lr){2-4} \cmidrule(lr){6-11}
& \multirow{2}{*}{Prompt Type} & \multirow{2}{*}{Prompt Family} & \multirow{2}{*}{Variant} 
&  
& \multicolumn{2}{c}{Perception (Q1)}
& \multicolumn{2}{c}{Prediction (Q2)}
& \multicolumn{2}{c}{Reasoning (Q3)} \\
\cmidrule(lr){6-7} \cmidrule(lr){8-9} \cmidrule(lr){10-11}
&  &  & 
&  
& BERT-F1 & SBERT
& BERT-F1 & SBERT
& BERT-F1 & SBERT \\
\midrule

\multirow{11}{*}{Phi-2-2.7B~\cite{javaheripi2023phi}}
& No System Prompt & -- & --& 0.463 & 0.365  &0.487& 0.366 &0.482  &  0.451 & 0.729 \\
\cline{2-11}
& \multirow{10}{*}{Textual System Prompts} & -- & Raw & 0.444 & 0.341 &0.453& 0.340 &0.462  &  0.433 & 0.713 \\
\cline{3-11}
&  & \multirow{3}{*}{A-Series} & A1 & 0.400 &0.347 & 0.564 &0.342  & 0.573 & 0.386 & 0.731 \\
&  &  & A2 & 0.344 & 0.345 &0.553 & 0.367 & 0.577 &0.373 &  0.723\\
&  &  & A3 & \textbf{0.481} & 0.395 &0.576  & 0.398 & 0.592 & 0.445 & 0.720\\
\cline{3-11}
&  & \multirow{3}{*}{R-Series} & R1 & 0.400 & 0.394 & 0.569 & 0.393 &  0.599 & 0.420 & 0.743 \\
&  &  & R2 & 0.388 &0.335  &0.537  &0.374  &0.556  &0.412  & 0.733 \\
&  &  & R3 & \textbf{0.463} & 0.338 & 0.522 & 0.356 &0.540  & 0.365 & 0.690 \\
\cline{3-11}
&  & \multirow{3}{*}{PR-Series} & PR1 & 0.381 & 0.321 &0.536  &0.349  & 0.545 &0.293  & 0.642 \\
&  &  & PR2 & 0.475 & 0.370 & 0.548 &  0.374 & 0.545 &0.391  & 0.659 \\
&  &  & PR3 & \textbf{0.513} & 0.398 & 0.592 &  0.401 & 0.598 &0.456  & 0.745 \\
\midrule

\multirow{11}{*}{Stablelm-2-zephyr-1\_6b~\cite{bellagente2024stable}}
& No System Prompt & -- & -- & 0.475 & 0.510 & 0.683 &0.465  &0.670  & 0.560 & 0.732 \\
\cline{2-11}
& \multirow{10}{*}{Textual System Prompts} & -- & Raw & 0.344 &  0.374 & 0.596 & 0.377 & 0.606 & 0.413 & 0.728 \\
\cline{3-11}
&  & \multirow{3}{*}{A-Series} & A1 & 0.450 & 0.390 & 0.569 & 0.370 & 0.573 & 0.442 & 0.689 \\
&  &  & A2 & 0.438 &0.374  &0.591  & 0.386 & 0.600 & 0.441 & 0.741 \\
&  &  & A3 & \textbf{0.469} &0.396  &0.557  & 0.376 & 0.577 & 0.450 & 0.743 \\
\cline{3-11}
&  & \multirow{3}{*}{R-Series} & R1 &  0.450&  0.372 & 0.502 &0.350   & 0.521 & 0.437 & 0.711 \\
&  &  & R2 & 0.425 &  0.403 &  0.584 & 0.414 &0.606  & 0.450 &0.727 \\
&  &  & R3 & \textbf{0.488} & 0.384 &  0.571 & 0.382 &0.588 &  0.450& 0.743 \\
\cline{3-11}
&  & \multirow{3}{*}{PR-Series} & PR1 &  0.319& 0.308 & 0.564 & 0.290 & 0.544 & 0.403 & 0.706 \\
&  &  & PR2 & 0.331 &  0.345 & 0.586 & 0.354 & 0.593 & 0.406 &0.716  \\
&  &  & PR3 & \textbf{0.494} & 0.513 & 0.691 & 0.486 & 0.696 & 0.570 &0.742 \\
\midrule

\multirow{11}{*}{TinyLlama-1.1B-Chat-v1.0~\cite{zhang2024tinyllama}}
& No System Prompt & -- & -- & 0.375 & 0.361 & 0.609 &  0.375 & 0.600 &0.454  & 0.766 \\
\cline{2-11}
& \multirow{10}{*}{Textual System Prompts} & -- & Raw & 0.344 & 0.337 & 0.583&  0.361& 0.585 &0.416  & 0.725 \\
\cline{3-11}
&  & \multirow{3}{*}{A-Series} & A1 & 0.381 &  0.298 & 0.568 & 0.360 &0.564  & 0.416 & 0.685 \\
&  &  & A2 & 0.419  & 0.383 & 0.611 & 0.397 &  0.610 & 0.452 &0.748 \\
&  &  & A3 & \textbf{0.450}  & 0.428 & 0.642 &  0.440 & 0.620 & 0.485 & 0.744 \\
\cline{3-11}
&  & \multirow{3}{*}{R-Series} & R1 & 0.538  & 0.447 & 0.651 & 0.433 &  0.634& 0.494 & 0.764 \\
&  &  & R2 & 0.531 & 0.457 &0.672   &0.448   &0.650  & 0.513 & 0.771 \\
&  &  & R3 & \textbf{0.544} & 0.458 &  0.691 & 0.452 &0.668  & 0.528 &0.797  \\
\cline{3-11}
&  & \multirow{3}{*}{PR-Series} & PR1 & 0.425 & 0.467 & 0.653 &0.442  &0.659  &0.488  & 0.743 \\
&  &  & PR2 & 0.269 & 0.447 & 0.654 & 0.448 & 0.643 & 0.490 & 0.759 \\
&  &  & PR3 & \textbf{0.563} & 0.497 & 0.686 &0.475  &0.663  &0.490  & 0.761 \\
\bottomrule
\end{tabular}
}
\end{table*}

\begin{figure*}[htbp]
    \centering
    \includegraphics[width=\linewidth]{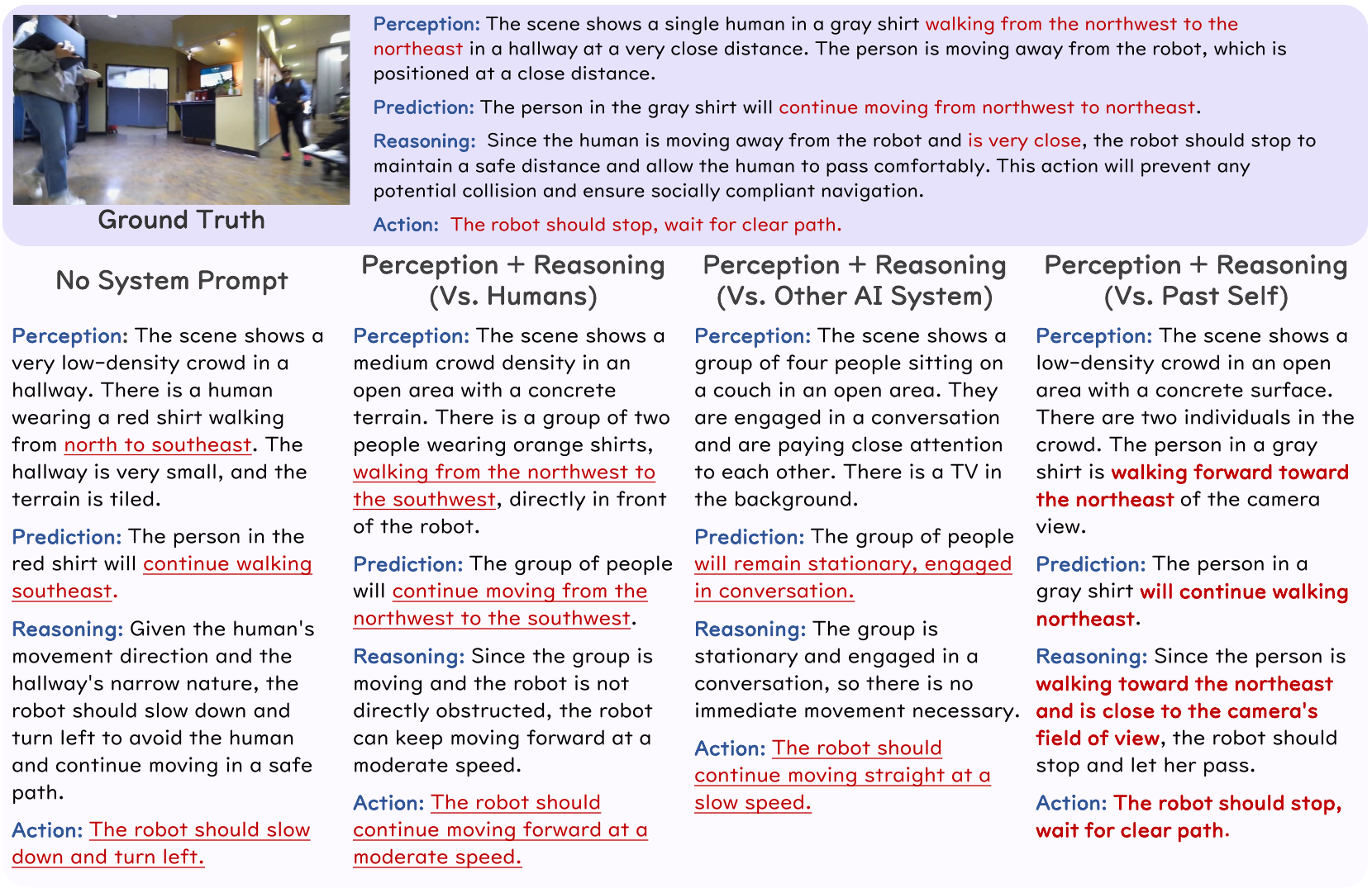}
    \caption{Visualization of perception, prediction, reasoning, and decision making for the finetuned model using TinyLlama-1.1B-Chat-v1.0 as the language model. The results indicate that prompts integrating perception and reasoning with competition against the model’s past behavior achieve the best overall performance.}
    \label{fig:vis}
\vspace{0.2cm}
    \centering
    \includegraphics[width=\linewidth]{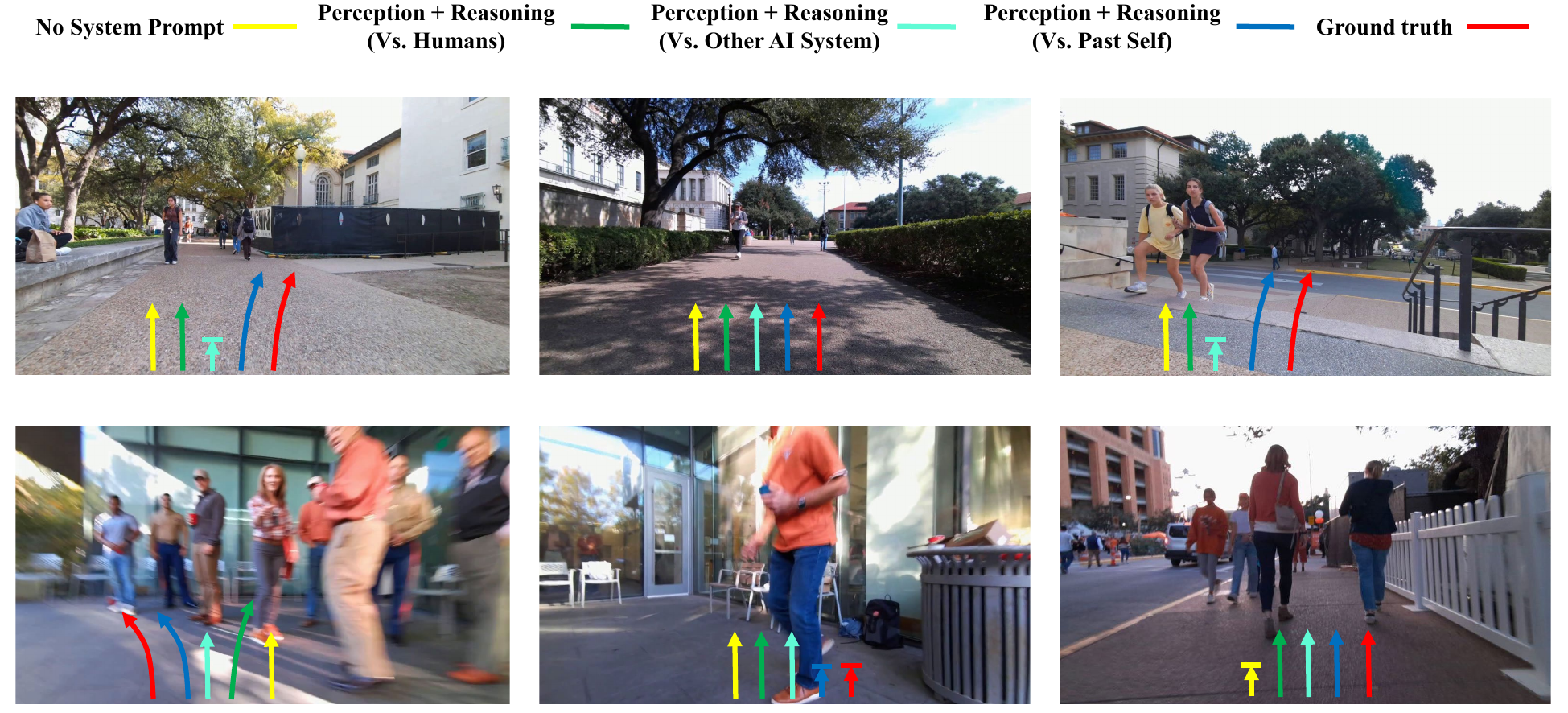}
    \caption{Visualization of predicted actions using TinyLlama 1.1B Chat v1.0 as the language model. In relatively easy scenarios, for example the middle image in the top row, all prompt settings produce correct decisions. In more challenging cases, such as crowded scenes or blurry observations, the prompt that integrates perception and reasoning and frames competition against the model’s past self yields more accurate and safer action decisions.}
    \label{fig:action}
\end{figure*}

\subsection{Evaluation Metrics.}
\noindent\textbf{Semantic Alignment (token-level)}: BERT-F1.
It measures semantic similarity between a generated prediction and a reference text using contextualized token embeddings.
Let the ground truth text be $y_t^{\text{gt}}=\{t_1,\dots,t_m\}$ and the predicted text be $y_t^{\text{pred}}=\{p_1,\dots,p_n\}$, where each token is represented by a contextual embedding.
Token level similarity is computed using cosine similarity between embeddings.
BERT-F1 is defined as
\begin{equation}
\begin{aligned}
\text{BERT-F1} &=
\frac{2 \cdot \text{BERT-P} \cdot \text{BERT-R}}
{\text{BERT-P} + \text{BERT-R}}, \\
\text{BERT-P} &=
\frac{1}{n} \sum_{i=1}^{n} \max_{1 \le j \le m} \cos(p_i, t_j), \\
\text{BERT-R} &=
\frac{1}{m} \sum_{j=1}^{m} \max_{1 \le i \le n} \cos(t_j, p_i).
\end{aligned}
\end{equation}
Here, $\text{BERT-P}$ measures the extent to which the generated content is supported by the reference, penalizing irrelevant or hallucinated additions, while $\text{BERT-R}$ measures how well the generation covers key information in the reference, penalizing omissions.
The $\text{BERT-F1}$ score provides a balanced assessment by jointly considering precision and recall.

\medskip
\noindent\textbf{Semantic Alignment (sentence-level)}: SBERT. It computes semantic closeness between a prediction and the ground truth using sentence level embeddings produced by Sentence BERT~\cite{reimers2019sentence}.
Let $\mathbf{y}_s^{\text{pred}}$ and $\mathbf{y}_s^{\text{gt}}$ denote the sentence embeddings of the prediction and ground truth, respectively.
The similarity is defined as
\begin{equation}
\text{SBERT} =
\cos(\mathbf{y}_s^{\text{pred}}, \mathbf{y}_s^{\text{gt}})
= \frac{\mathbf{y}_s^{\text{pred}} \cdot \mathbf{y}_s^{\text{gt}}}
{\|\mathbf{y}_s^{\text{pred}}\| \, \|\mathbf{y}_s^{\text{gt}}\|}.
\end{equation}

\medskip
\noindent\textbf{Action Safety}: Action Accuracy (AA).
This score evaluates decision level correctness by assessing whether the robot action inferred from the model output matches the ground truth action.
Let $y_c^{\text{pred}}$ and $y_c^{\text{gt}}$ denote the predicted and ground truth action labels for the $c$ th sample, and let $N$ denote the total number of samples.
Action Accuracy is defined as
\begin{equation}
\text{AA} = \frac{1}{N} \sum_{c=1}^{N}
\mathbb{I}\!\left(y_c^{\text{pred}} = y_c^{\text{gt}}\right),
\end{equation}
where $\mathbb{I}(\cdot)$ is the indicator function.
Unlike semantic similarity metrics, $\text{AA}$ directly reflects whether the model selects the correct control decision.

\section{Experimental Results}
\subsection{Main Results.}
\noindent\textbf{Non-finetuned Model.}
We first investigate the effectiveness of the proposed prompt engineering strategies in a zero shot VLM setting. Since GPT-4o currently represents one of the strongest general purpose VLMs, we adopt it as the evaluation model for this analysis.
To ensure a fair and consistent comparison across different prompt designs, we constrain the model action space to a fixed set of discrete navigation commands. This design choice enables direct and unambiguous evaluation of the generated actions while preserving the model internal reasoning process. Specifically, the allowable actions include ``Move Forward'', ``Forward Left'', ``Forward Right'', ``Stop'', ``Turn Left'', and ``Turn Right''. The results are summarized in Table~\ref{tab:gpt-4o}. The ``No System Prompt'' setting indicates that no prefixed instruction is provided and the model receives only the input image and the user query without additional guidance or task framing. This configuration evaluates the inherent ability of GPT-4o to perform socially compliant navigation in the absence of explicit prompting.

Overall, for non-finetuned GPT-4o, competition against humans yields the best performance, whereas competition against other AI models results in the worst performance. However, performance is also influenced by the interaction among prompt design, the language model, and the dataset.
Interestingly, across both datasets, a naive system prompt performs worse than the no–system-prompt baseline, indicating that the use of system prompts is not universally beneficial. This finding highlights that prompt engineering must be carefully designed, as inappropriate or weakly aligned prompts can degrade model performance rather than improve it.

\noindent\textbf{Finetuned Small VLMs.}
We further evaluate finetuned small VLMs, with results summarized in Tables~\ref{tab:SNEI_AA_Semantic} and~\ref{tab:MUSON_AA_Semantic}. Across different language models, competition against the model’s past self consistently yields the best performance, followed by competition against humans, while competition against other AI systems remains the least effective. Nevertheless, performance is still influenced by residual coupling effects among datasets, model variants, and prompt designs. Notably, the proposed system prompts lead to substantially larger improvements in action accuracy than in the semantic quality of generated language responses. This suggests that the proposed prompts mainly constrains the final action selection process, improving the alignment between model reasoning and discrete action decisions, rather than broadly enhancing semantic expressiveness.

\subsection{Visualizations.}
As shown in Fig.~\ref{fig:vis}, we qualitatively compare perception, prediction, reasoning, and decision-making behaviors of a finetuned TinyLlama-1.1B-Chat-v1.0 model under different prompt strategies on the SNEI dataset. Without a system prompt, the model often exhibits weakly grounded reasoning, leading to unsafe navigation actions and potential human collisions. In contrast, prompts that integrate perception and reasoning and encourage competition against the model’s past behavior yield more accurate scene understanding, more coherent human motion predictions, and safer decisions. Fig.~\ref{fig:action} further visualizes the resulting action predictions under different prompt settings.

\section{Limitations and Future Work}
First, while the datasets used in this study enable controlled evaluation, their limited size may not fully capture the diversity of real-world social navigation scenarios, such as snowy weather, nighttime conditions, or situations involving traffic signals. Future work will consider larger and more diverse datasets to better assess generalization.
Then, we focus on text-based prompt design and do not explore adaptive or learned prompting mechanisms. Incorporating online prompt adaptation and reinforcement learning to respond to environmental feedback is an important direction for future research.
Finally, our experiments assume a constrained, discrete action space, which simplifies comparison but limits realism. Extending our analysis to continuous control and more complex embodied environments remains future work.

\section{Conclusions}
In this paper, we study prompt design for socially compliant navigation in efficiency-oriented settings based on small VLMs. Inspired by theories of human cognitive learning and behavioral motivation, we investigate system guidance and motivational framing as two key dimensions of prompt design.
Experiments on the SNEI and MUSON datasets show that optimal prompt configurations differ across models. For zero-shot GPT-4o, reasoning-oriented prompts with competition against humans achieve the best performance, whereas for finetuned small VLMs, prompts that integrate perception and reasoning combined with competition against the model’s past behavior perform best. Nevertheless, action accuracy is also influenced by coupling effects among prompt design, model choice, and dataset characteristics, leading to occasional exceptions.
We further show that system prompts are not universally beneficial: poorly designed prompts can underperform a no–system-prompt baseline. Moreover, while finetuning primarily improves semantic-level understanding, our proposed prompts yield larger gains in action accuracy, indicating that prompt design mainly acts as a decision-level constraint.
Overall, our results demonstrate that principled, motivation-aware prompt design can substantially improve decision accuracy and social compliance without increasing model size, offering practical guidance for deploying small VLMs in real-world robot navigation.

\section*{Acknowledgments}
This paper is financially supported by JSPS KAKENHI Grant Number 24K20787.

\bibliographystyle{named}
\bibliography{ijcai26}

\end{document}